\documentclass[11pt,a4paper]{article}
\usepackage[hyperref]{acl2021}
\usepackage{times}
\usepackage{latexsym}

\usepackage{microtype}

\aclfinalcopy 

\usepackage{color}
\usepackage{kotex}
\usepackage{comment}
\usepackage{amsmath,amssymb,amsfonts}
\usepackage{mathtools}
\usepackage{tabulary}
\usepackage{multirow}
\usepackage{pifont}
\usepackage{algorithm}
\usepackage{algorithmic}    
\usepackage{xspace}
\newcommand{\xmark}{\ding{55}}

\newcommand{\rv}[1]{{\textcolor{black}{#1}}}

\newcommand{\toolnameMeta}{MetaUMT\xspace}
\newcommand{\ourtoolname}{MetaGUMT\xspace}

\usepackage{chngcntr}
\usepackage{amsmath}

\title{Unsupervised Neural Machine Translation for Low-Resource Domains via Meta-Learning}

\author{ Cheonbok Park$^{1,*}$, Yunwon Tae$^{2,*}$,Taehee Kim$^3$, Soyoung Yang$^3$, \\ \textbf{Mohammad Azam Khan$^3$ 
Lucy Park$^4$, and Jaegul Choo$^3$} \\
    $^1$NAVER,cbok.park@navercorp.com, \\
  $^2$Korea University, tyj204@korea.ac.kr\\ 
  $^3$KAIST,~\{taeheekim, sy\_yang, azamkhan, jchoo\}@kaist.com\\
  $^4$Upstage AI Research, lucy@upstage.ai
  }
\date{}
\begin{document}
\maketitle
\begin{abstract}
Unsupervised machine translation, which utilizes unpaired monolingual corpora as training data, has achieved comparable performance against supervised machine translation. However, it still suffers from data-scarce domains. To address this issue, this paper presents a novel meta-learning algorithm for unsupervised neural machine translation (UNMT) that trains the model to adapt to another domain by utilizing only a small amount of training data. We assume that domain-general knowledge is a significant factor in handling data-scarce domains. Hence, we extend the meta-learning algorithm, which utilizes knowledge learned from high-resource domains, to boost the performance of low-resource UNMT. Our model surpasses a transfer learning-based approach by up to 2-4 BLEU scores. Extensive experimental results show that our proposed algorithm is pertinent for fast adaptation and consistently outperforms other baseline models.
\end{abstract}
\let\thefootnote\relax\footnotetext{$*$ equal contributions}
\section{Introduction}

Unsupervised neural machine translation (UNMT) leverages unpaired monolingual corpora for its training, without requiring an already labeled, parallel corpus. 
Recently, the state of the art in UNMT~\cite{conneau2019cross,song2019mass,ren2019explicit} has achieved comparable performances against supervised neural machine translation (NMT) approaches. 
Instead of not using parallel corpus, training the UNMT model requires a significant amount of monolingual sentences (e.g., 1M-3M sentences). However, the prerequisite limits UNMT's applicability to low-resource domains, especially for the domain-specific document translation tasks. 
Since those documents require their knowledge for translation, the monolingual data themselves are scarce and expensive. 



Yet, UNMT for low-resource domains is not an actively explored field. One naive approach is to train a model on high-resource domains (e.g., economy and sports) while hoping the model will generalize on an unseen low-resource domain (e.g., medicine). However, recent studies have shown that non-trivial domain mismatch can significantly cause low translation accuracy on supervised NMT tasks~\cite{koehn2017six}.



Another reasonable approach is transfer learning, particularly, domain adaptation, which has shown performance improvements in the supervised NMT literature~\cite{freitag2016fast,zeng2019iterative}. In this approach, the model is first pretrained using data from existing domains and then finetuned by a new domain. However, this approach can suffer from overfitting and catastrophic forgetting due to a small amount of training data and a large domain gap. 





As an effective method for handling a small amount of training data, meta-learning has shown its superiority in various NLP studies such as dialog generation, machine translation, and natural language understanding~\cite{qian2019domain,gu2018meta,dou2019investigating}. \rv{In general, the meta-learning approach is strongly affected by the number of different tasks where tasks are defined as languages or domains from the aforementioned studies. However, in practice, previous studies may struggle to gather data to define tasks because they rely on a supervised model that requires labeled corpora. In this respect, we argue that applying a meta-learning approach to the unsupervised model is more feasible and achievable than the supervised model because it can define multiple different tasks with unlabeled corpora. Therefore, we newly introduce a meta-learning approach for UNMT, called MetaUMT, for low-resource domains by defining each task as a domain.}


\begin{figure*}[]
    \centering
    \includegraphics[width=1\textwidth]{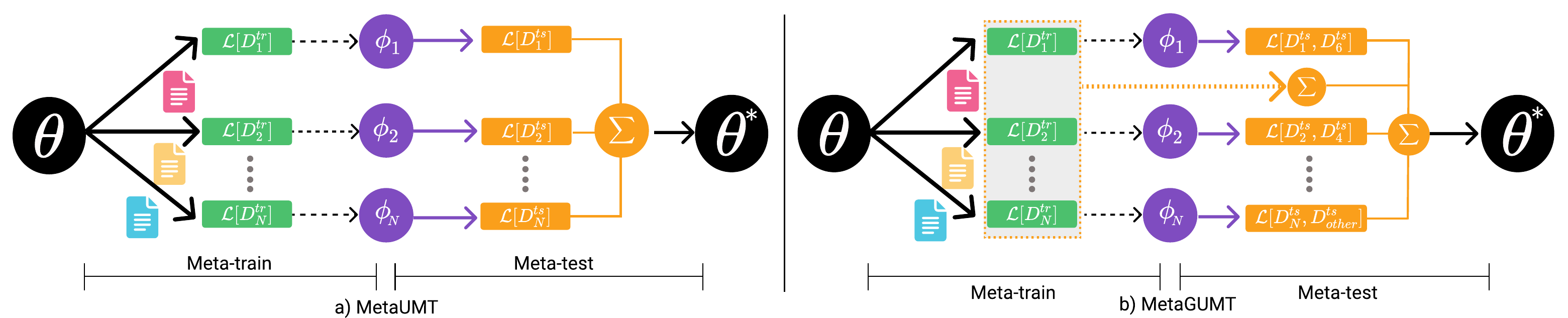}
    \caption{An illustration of high-level training process for both MetaUMT and MetaGUMT. In the case of MetaUMT, the training process is divided into two different phases, a meta-train phase and a meta-test phase. In the meta-train phase, $\theta$ adapts to a specific domain by leveraging the meta-train loss, i.e., $\mathcal{L}[D^{tr}_{N}]$, to obtain adapted parameters, i.e., $\phi_{N}$. $N$ represents the number of domains; $tr$ indicates meta-train data. In the meta-test phase, we optimize initial parameters $\theta$ through $\phi$ using the meta-test losses, i.e., $\sum{\mathcal{L}[D^{ts}_{N}]}$, where $ts$ indicates meta-test data.} 
    \label{fig.algo_overview}
    \vspace{-0.2in}
\end{figure*}





\rv{The objective of \toolnameMeta is to find the optimal initialization of the model parameters for neural machine translation that can quickly adapt to a new domain even with only a small amount of monolingual data. As shown in Fig.~\ref{fig.algo_overview} (a), we define two different training phases, i.e., a meta-train and a meta-test phase, and simulate the domain adaption process in order to obtain optimal initial parameters.
To be specific, the meta-train phase adapts to a domain while the meta-test phase optimizes initial parameters by leveraging adapted parameters obtained from the meta-train phase. After obtaining optimal initial parameters through these two phases, we fine-tune the model using a target domain, i.e., low-resource domain.}


\rv{Although the initial parameters optimized through \toolnameMeta are suited to adapt to the low-resource domain, these parameters may not fully maintain the knowledge of high-resource domains. For instance, in the meta-test phase, \toolnameMeta optimizes initial parameters using the adapted parameters; however, it discards meta-train knowledge used to update adapted parameters in the meta-train phase. Moreover, instead of validating the same domain used in the meta-train phase, we hope to inject generalizable knowledge into initial parameters by utilizing another domain in the meta-test phase. This prevents overfitting from the data scarcity issue.}

\rv{As shown in Fig.~\ref{fig.algo_overview} (b), we propose an improved meta-learning approach called \ourtoolname for low-resource UNMT by explicitly promoting common knowledge across multiple source domains as well as generalizable knowledge from one particular domain to another. In other words, we do not only encourage the model to find the optimal initial parameters that can quickly adapt to a target domain with low-resource data, but also encourage the model to maintain common knowledge, e.g., general words such as determiners, conjunctions, and pronouns, which is applicable from multiple source domains. Furthermore, due to a small number of training data in a low-resource domain, the model can suffer from overfitting; however, we attempt to handle overfitting by leveraging generalizable knowledge that is available from one domain to another. Our proposed meta-learning approach demonstrates a consistent efficacy against other baseline models.}

Overall, our contributions can be summarized as follows: (1) We apply a meta-learning approach for UNMT. To the best of our knowledge, this is the first study to use a meta-learning approach for UNMT, where this approach is more suitable to a UNMT task than a supervised one; (2) We empirically demonstrate that our enhanced method, \ourtoolname, shows fast convergence on both pre-training (i.e., meta-learning with source domains) and finetuning (i.e., adapting to a target domain); (3) The model trained with \ourtoolname consistently outperforms all baseline models including \toolnameMeta. This demonstrates that finding optimal initial parameters that incorporate high-resource domain knowledge and generalizable knowledge is significant to handle a low-resource domain.

\section{Related Work}
Our study leverages two components from the natural language processing (NLP) domain: low-resource NMT and meta-learning. 
In this section, we discuss previous studies by concentrating on the main components.

\subsection{Low-Resource Neural Machine Translation}
Based on the success of attention-based models~\cite{luong2015effective,vaswani2017attention}, NMT obtain significant improvement in numerous language datasets, even showing human-like performances~\cite{wu2016google} in different datasets. However, the performance of NMT models depends on a size of parallel sentences from the source and target language~\cite{koehn2017six}. To address this problem, diverse approaches have been proposed, which are categorized into two different directions: (1) utilizing monolingual datasets and (2) transferring the knowledge from high-resource domains to a low-resource domain. 
 
Recent studies point out the difficulty of gathering the parallel data
, whereas the monolingual datasets are relatively easy to collect. 
To facilitate the monolingual corpora, several studies apply dual learning~\cite{he2016dual}, back-translation~\cite{sennrich2016neural}, and pretraining the model with the bilingual corpora~\cite{hu-etal-2019-domain-adaptation,wei2020iterative}.
Furthermore, as a challenging scenario, recent studies propose the UNMT methods without using any parallel corpus~\cite{lample2017unsupervised,artetxe2017unsupervised,yang2018unsupervised}. 
The UNMT models show comparable performances by extending the back-translation method~\cite{conneau2017word} and incorporating the methods for good initialization, such as the shared byte pair encoding (BPE)~\cite{lample2018phrase} and the cross-lingual representations~\cite{conneau2019cross}, following the ones of the supervised NMT.
However, theses approaches still require plenty of monolingual or parallel dataset to adapt the model to the target domain.
Secondly, a few studies concentrate on transferring the knowledge from the rich-resources corpora into the low-resource one.
Several models~\cite{chu2018survey,hu2019domain} show better performances than when trained with the low-resource corpora only. 
Despite the improvements by the transfer learning approaches,
these approaches apply in constraint conditions, which are one or both of target domains or source domain corpus are the parallel corpus. For example, if we intend to create a translation system of a particular language in a particular domain, there may be fewer sentences in-domain as far as parallel out-of-domain data is scarce.

To address the issues, we define a new task as the unsupervised domain adaptation on the low-resource dataset.
Our work is a more challenging one than any other previous studies, \rv{since we assume that both the low-resource target domain and the source domain corpora are monolingual.}

\subsection{Meta Learning}

Given a small amount of training data, most of machine learning models are prone to overfitting, thus failing to find a generalizable solution. 
To handle this issue, meta-learning approaches seek for how to adapt quickly and accurately to a low-resource task, and show impressive results in various domains~\cite{finn2017model,javed2019meta}. 
\rv{The meta-learning approaches aim to find the optimal initialization of the model parameters which adapts the model to the low-resource dataset in a few iterations of training~\cite{finn2017model,ravi2016optimization}. 
Owing to the success of the meta learning, recent studies apply the meta-learning to the low-resource NMT tasks, including multi-lingual NMT~\cite{gu2018meta} and the domain adaptation~\cite{li2020metamt}. 
These studies assume that all the training corpora consist of the parallel sentences.} 
\rv{On the other hand, a recent work~\cite{li2017learning} utilizes the meta learning approach to find a generalized model for multiple target tasks. However, it is not focused on adapting a specific target task since its main goal is to handle the target task without using any low-resource data.}

\rv{Our study attempts to address the low-resource UNMT by exploiting meta-learning approaches.  Moreover, we present two novel losses that encourage incorporating high-resource knowledge and generalizable knowledge into the optimal initial parameters. Our proposed approaches show significant performance improvements in adapting to a low-resource target domain.}

\begin{figure*}[t]
    \centering
    \includegraphics[width=1\textwidth]{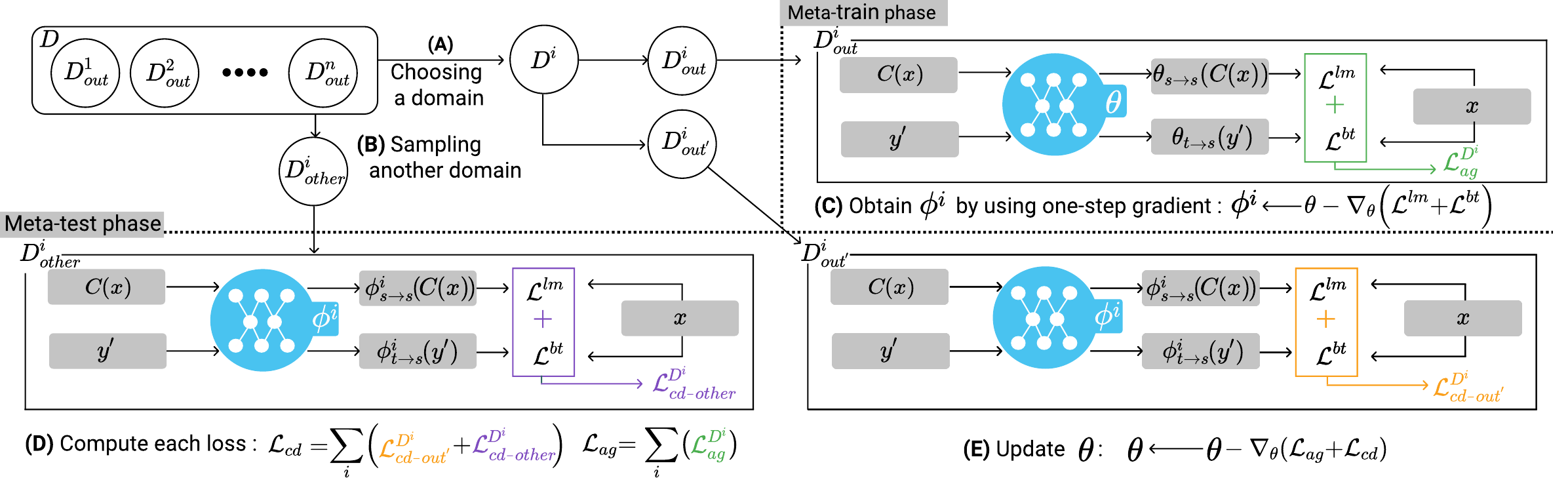}
    \caption{\rv{Overall training process of our proposed MetaGUMT. \textbf{(A)} A single domain (e.g., Law) is first chosen to compute $\mathcal{L}^{\mathcal{D}^{i}}_{ag}$ with model parameters $\theta$ in the meta-train phase and $\mathcal{L}^{\mathcal{D}^{i}}_{cd-out^{\prime}}$ with temporary model parameters $\phi^{i}$ in the meta-test phase. \textbf{(B)} Another domain (e.g., IT) is sampled to compute $\mathcal{L}^{\mathcal{D}^{i}}_{cd-other}$ based on $\phi^{i}$ in the meta-test phase. \textbf{(C)} Temporary model parameters $\phi^{i}$ is updated from $\theta$ to learn the knowledge of high-resource domains. \textbf{(D)} Cross-domain and aggregated meta-train loss functions are computed across all out-domain datasets. \textbf{(E)} The optimal initialization $\theta$ is obtained by minimizing $\mathcal{L}_{ag}$ and $\mathcal{L}_{cd}$.}}
    \label{fig:fig_algo}
    \vspace{-3mm}
\end{figure*}

\section{Unsupervised Neural Machine Translation}\label{section.2}
In this section, we first introduce the notation of the general UNMT models. We then describe the three steps for the UNMT task:
initialization, language modeling, and back-translation. On these three steps, we illustrate how each step contributes to improving the performance of UNMT.

\textbf{Notations}.
We denote $S$ and $T$ as a source and a target monolingual language datasets. $x$ and $y$ represent the source and the target sentences from $S$ and $T$. 
We assume the NMT model is parameterized by $\theta$. We also denote $M_{\text{s}\rightarrow\text{s}}$ and $M_{\text{t}\rightarrow\text{t}}$ as language models in a source and a target languages, respectively, while denoting $M_{\text{s}\rightarrow\text{t}}$ and $M_{\text{t}\rightarrow\text{s}}$  as the machine translation models from the source to the target language, and vice versa.

\textbf{Initialization}.
A recent UNMT model~\cite{lample2018phrase} is based on a shared encoder and decoder architecture for the source and the target language. 
Due to the shared encoder and decoder for each language, initializing the model parameters of the shared encoder and decoder is an important step for competitive performances~\cite{conneau2017word,lample2017unsupervised,artetxe2017unsupervised,yang2018unsupervised}. \citet{conneau2019cross} propose the XLM (cross-lingual language model) to initialize parameters, showing the significantly improved performances for UNMT. 
Among various initialization methods, we leverage the XLM as our initialization method.

\textbf{Language modeling}.
We use a denoising auto-encoder~\cite{vincent2008extracting} to train the UNMT model, reconstructing an original sentence from a noisy one in a given language. The objective function is defined as follows:

\begin{equation}\label{eqn:LM}
\begin{split}
\mathcal{L^{\text{lm}}}  = & \mathbb{E}_{x \sim S}[-\log M_{s\rightarrow s}(x|C(x))] + \\
& \mathbb{E}_{y \sim T}[-\log M_{t\rightarrow t}(y|C(y))], 
\end{split}
\end{equation}

where $C$ is a noise function described in \cite{lample2018phrase}, which randomly drops or swaps words in a given sentence.
By reconstructing the sentence from the noisy sentence, the model learns the language modeling in each language.

\textbf{Back-translation}.
\rv{Back-translation helps the model learn the mapping functions between the source and the target language by using only those monolingual sentences.}
For example, we sample a sentence $x$ and $y$ from source language $S$ and target language $T$. To make pseudo-pair sentences from the sampled source sentence, we deduce the target sentence from the source sentence, such that $ y^{\prime}=M_{\text{s}\rightarrow\text{t}}\left(x\right)$. Finally, we get the pseudo parallel sentence, i.e., $\left(x,y^{\prime}\right)$. Similarly, we obtain $\left(x^{\prime},y\right)$, where $x^{\prime}$ is the translation of a target sentence, i.e., $M_{\text{t}\rightarrow\text{s}}\left(y\right)$.
We do not back-propagate when we generate the pseudo-parallel sentence pairs.
In short, the back-translation objective function is
\begin{equation}\label{eqn:BT}
\begin{split}
\mathcal{L^{\text{bt}}} =& \mathbb{E}_{y\sim {T}} [-log {M}_{s\rightarrow t}\left(y\mid x^{\prime} \right)] +\\
& \mathbb{E}_{x\sim {S}} [-log {M}_{t\rightarrow s}\left(x\mid y^{\prime} \right)].
\end{split}  
\end{equation}




%

\section{Proposed Approach}
This section first explains our formulation of a low-resource unsupervised machine translation task where we can apply a meta-learning approach. Afterwards, we elaborate our proposed methods, \toolnameMeta and \ourtoolname. We utilize the meta-learning approach to address a low-resource challenge for unsupervised machine translation. Moreover, we extend \toolnameMeta into \ourtoolname to explicitly incorporate learned knowledge from multiple domains.

\subsection{Problem Setup}
\citet{finn2017model} assume multiple different tasks to find the proper initial parameters that can quickly adapt to a new task using only a few training examples. In this paper, we consider tasks in the meta-learning as domains, where $\mathcal{D}_{out } = \{\mathcal{D}_{out }^0,...,\mathcal{D}_{out }^{n}\}$ represents $n$ out-domain datasets (i.e., source domain datasets), and $\mathcal{D}_{in}$ indicates an in-domain dataset (i.e., a target domain dataset), which can be the dataset in an arbitrary domain not included in $\mathcal{D}_{out}$. Each domain in both $\mathcal{D}_{out}$ and $\mathcal{D}_{in}$ is assumed to be composed of unpaired language corpora, and we create $\mathcal{D}_{in}$ as a low-resource monolingual dataset~\footnote{We randomly sample the 5,000 words ($\sim 300$ sentences) from the in-domain training dataset.}. To adapt our model to the low-resource in-domain data, we finetune the UNMT model by minimizing both the losses described in Eqs.~\eqref{eqn:LM} and \eqref{eqn:BT} with $\mathcal{D}_{in}$.



\subsection{{\toolnameMeta}}
In order to obtain an optimal initialization of model parameters, allowing the model to quickly adapt to the new domain with only a small number of monolingual training data, \toolnameMeta uses two training phases, the \textit{meta-train} phase and the \textit{meta-test} phase. 
During the meta-train phase, the model first learns a domain-specific knowledge by updating initial model parameters $\theta$ to temporary model parameters $\phi^{i}$, i.e., adapted parameters. Then, in the meta-test phase, the model learns the adaptation by optimizing $\theta$ with respect to $\phi^{i}$. From the domain adaption perspective, two phases simulate the domain adaption process. The model first adapts to a specific domain through the meta-train phase, and this adaption is evaluated in the meta-test phase.






\textbf{Meta-train phase}. We obtain $\phi^{i}$ for each i-$th$ out-domain datasets by using one-step gradient descent, i.e., 
\begin{equation}\label{eqn:meta_train.1}
\begin{array}{r@{}l}
\phi^{i} &= \theta - \alpha\nabla_{\theta}\mathcal{L}^{\text{s}}_{\mathcal{D}^\text{i}_{out} }(\theta),
\end{array}
\end{equation}
where $\mathcal{L}^{s}_{\mathcal{D}^\text{i}_{out} }$ is represented as
\begin{equation}\label{eqn:meta_train.2}
\begin{array}{r@{}l}
    \mathcal{L}^{s}_{\mathcal{D}^\text{i}_{out} } &=
    \mathcal{L}^{\text{lm}}_{{\mathcal{D}}^{\text{i}}_{out }}(\theta)+\mathcal{L}^{\text{bt}}_{{\mathcal{D}}^{\text{i}}_{out }}(\theta).
\end{array}
\end{equation}
$\mathcal{D}^{i}_{out }$ is the $i$-th out-domain dataset and $\alpha$ is the learning rate for the meta-train phase. As previously discussed in Section~\ref{section.2}, the language modeling and back-translation losses are essential in facilitating the unsupervised machine translation. 
Hence, $\mathcal{L}^{s}$ consists of $\mathcal{L^{\text{lm}}}$ and $\mathcal{L^{\text{bt}}}$, where each loss function is computed with $\mathcal{D}^{i}_{out}$.


 

\textbf{Meta-test phase}.
The objective of the meta-test phase is to update $\theta$ using each $\phi^{i}$ learned from the meta-train phase by using each $\mathcal{L}^{\text{s}}_{\mathcal{D}^{i}_{out' }}$~\footnote{$\mathcal{L}^{\text{s}}_{\mathcal{D}^{i}_{out}}$ and $\mathcal{L}^{\text{s}}_{\mathcal{D}^{i}_{out'}}$ indicate different batch sampled data from same $\mathcal{D}^{i}$.}.
We call this update as a meta-update, defined as
\begin{equation}\label{eqn:meta_test.2}
\theta \leftarrow \theta - \beta\nabla_{\theta}\sum^{n}_{i=0}\mathcal{L}^{\text{s}}_{\mathcal{D}^{i}_{out'}}(\phi^{i}),
\end{equation}
where $\beta$ is another learning rate in the meta-test phase. 
Since Eq.~\eqref{eqn:meta_test.2} requires the second-order gradient, the equation is simplified with the first-order gradient by replacing the second-order term. \citet{finn2017model} showed that the first-order approximation of the meta-learning maintains the performance while minimizing the computational cost. 

\subsection{{\ourtoolname}}
To handle a data scarcity issue from a meta-learning perspective, it is critical to be able to make the initialized model to adapt to a data-scarce domain. However, since a small amount of training data in the new domain may cause the model to overfit and prevent utilizing high-resource domain knowledge, it is important to incorporate high-resource domain knowledge and generalizable knowledge into optimal initial parameters. To address this issue, we extend the existing meta-learning approach via two novel losses, which we call \rv{an aggregated meta-train loss and a cross-domain loss}. \rv{The former contributes to incorporating those high-resource domain knowledge into optimal initial parameters, while the latter encourages our model, after trained using a particular domain, to still generalize well to another domain, i.e., cross-domain generalization.}

\textbf{Meta-train phase}.
As shown in Fig.~\ref{fig:fig_algo} (C), via Eqs.~\eqref{eqn:meta_train.1} and \eqref{eqn:meta_train.2}, we obtain $\phi^{i}$ from each i-$th$ out-domain datasets. Since this phase is exactly same with the meta-train phase of \toolnameMeta, we leave out the details.
%

\begin{table*}[t]
\centering
\resizebox{\textwidth}{!}{\begin{tabular}{l|r|rrr|rrr|rrr|rrr}
\hline
\multirow{4}{*}{Model}   &
    \multirow{2}{*}{$\mathcal{D}_{out}$} 
                    & Medical & Law & EUB & Medical & Law & EUB & Subtitles & Law &EUB&  GV & Europarl &EUB \\
                    & & Koran & IT & GV &  Koran & IT & GV & Europarl& IT& GV & Subtitles& Medical& Koran \\
                    
                    \cline{2-14}
    & \multirow{2}{*}{$\mathcal{D}_{in}$} 
                    & \multicolumn{3}{c|}{Subtitles} & \multicolumn{3}{c|}{Europarl} & \multicolumn{3}{c|}{Medical} & \multicolumn{3}{c}{Law}\\\cline{3-14}
                    & &  De-En & En-De & epoch & De-En & En-De & epoch & De-En & En-De & epoch &  De-En & En-De & epoch   \\
                    \hline
\multicolumn{2}{l|}{Unadapted}  &  9.46 & 7.54 & - & 22.31 & 15.82 & - & 21.3  & 19.23 & - & 31.1  & 25.35& - \\
\multicolumn{2}{l|}{Transfer} &10.92 &  9.18 & 4  & 22.96 & 16.78& 3 & 22.77 & 19.78 & 6 & 31.69 & 25.59 & 4  \\ 
\multicolumn{2}{l|}{Mixed} & 11.77 & 9.96& 15 & 22.99 & 17.05& 5 & 22.98 & 19.99 & 8 & 31.69 & 25.74 & 6  \\ 
\multicolumn{2}{l|}{{\toolnameMeta}} & 12.95& 10.58 & 3 & 24.53& 18.59& \textbf{2}& 24.6 & 21.86 & \textbf{4}&  32.51 & 27.22 & 3   \\ 
\multicolumn{2}{l|}{{\ourtoolname}} &  \textbf{13.45} &\textbf{10.89} & \textbf{2} &  \textbf{25.13}& \textbf{18.95}& \textbf{2}& \textbf{25.32}& \textbf{22.79} & \textbf{4}& \textbf{34.26}& \textbf{29.37} & \textbf{2}\\ 
\hline    
\multicolumn{2}{l|}{Supervised NMT} & 2.24 & 2.49 & 8 & 1.88 & 1.52 & 7 & 7.71 & 9.8 & 11 & 11.29 & 10.07 & 13 \\
\multicolumn{2}{l|}{Unsupervised NMT} & 1.26 & 0.94 & 5 & 1.53 & 0.76 & 23 & 3.37 & 2.72 & 9 & 6.07 & 4.73 & 11 \\
\hline
\end{tabular}}
\caption{Average BLEU scores on various out-domain ($D_{out}$) and in-domain ($D_{in}$) combinations for the language pairs of De-En and En-De. The `epoch' column indicates the converged number of epochs for each in-domain dataset. Since the unadapted model does not have any additional finetuning step, we leave the epoch column as blank. The bold represents the significant difference ($p < 0.05$) with others.}
\label{table.main_result}
\vspace{-0.2in}
\end{table*}
\textbf{Meta-test phase}.\label{section3.3}
The aggregated meta-train loss, which refers to Fig.~\ref{fig:fig_algo} (D), is computed using all out-domain datasets, i.e.,
\begin{equation}\label{eqn:meta_test.4}
\mathcal{L}_{ag}=\sum^{n}_{i=0}\mathcal{L}^{\text{s}}_{\mathcal{D}^{i}_{out}}(\theta). 
\end{equation}
This loss term allows the model to learn the source domain knowledge potentially applicable to a target domain. Moreover, \rv{to alleviate the overfitting after adapting to the low-resource domain}, we introduce a cross-domain loss, which is in Fig.~\ref{fig:fig_algo} (D), as 
\begin{equation}\label{eqn:meta_test.5}\mathcal{L}_{cd}=\sum^{n}_{i=0}\mathcal{L}^{\text{s}}_{\mathcal{D}^{i}_{cd}}(\phi^{i}),\end{equation}
where $\mathcal{L}^{\text{s}}_{\mathcal{D}^{i}_{cd}}=\mathcal{L}^{\text{s}}_{\mathcal{D}^{i}_{out'}}(\phi^{i}) + \mathcal{L}^{\text{s}}_{\mathcal{D}^{i}_{other}}(\phi^{i})$, i.e., computing the cross-domain loss with the data from ${\mathcal{D}^{i}_{out'}}$ as well as those from other domains than ${\mathcal{D}^{i}_{out'}}$ called $\mathcal{L}^{\text{s}}_{\mathcal{D}^{i}_{other}}$.




To obtain the optimal initialization $\theta$ for model parameters, we define our total loss function, which is Fig.~\ref{fig:fig_algo} (E), as the sum of the two of our losses, i.e.,
\begin{equation}\label{eqn:meta_test.3}
\theta \leftarrow \theta - \beta\nabla_{\theta}(\mathcal{L}_{cd} + \mathcal{L}_{ag}).
\end{equation}


In summary, our aggregated meta-train and cross-domain losses encourage our model to accurately and quickly adapt to an unseen target domain. The overall procedure is described in Algorithm~\ref{alg:our_maml}.

\section{Experiments}

This section first introduces experiment settings and training details. Afterwards, we show empirical results on 
various scenarios.
\subsection{Dataset and Preprocessing}


We conduct eight different domains~\footnote{Acquis (Law), EMEA (Medical), IT, Tanzil (Koran), Subtitles, EUbookshop (EUB), Europarl, and GlobalVoices (GV)} for our experiments (Appendix~\ref{table.dataset}). Each domain dataset is publicly available on OPUS~\footnote{http://opus.nlpl.eu/}~\cite{tiedemann2012parallel}. 
We utilize the eight different domains into out-domain ($D_{out}$) and in-domain datasets ($D_{in}$).
To build the monolingual corpora of in-domain and out-domain datasets, 
we divide each corpus into monolingual ones and shuffle these monolingual copra. 
Each of two monolingual corpora contains the equal number of sentences for each language (e.g., English and German). 
For our low-resource scenarios, we sample 5,000 tokens from a selected in-domain corpus for each language. 
Note that out-domain dataset represents full monolingual corpora. 

\begin{figure*}[t]
    \centering
    \includegraphics[width=\textwidth]{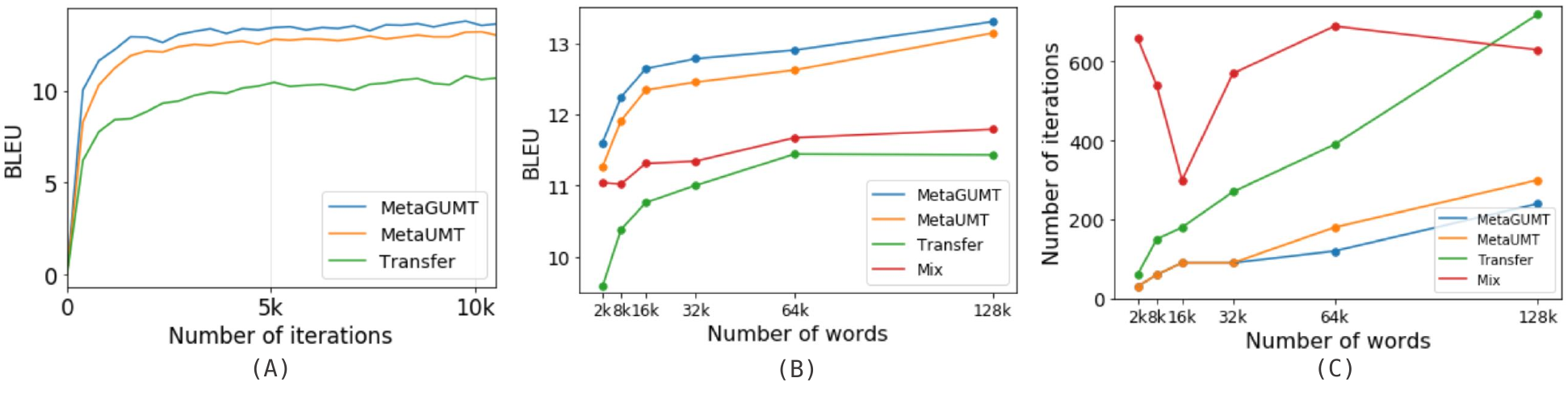}
    \caption{Results of the models that are first pretrained on Medical, Law, EUbookshop, Koran, IT, and GlobalVoices datasets and then finetuned on a Subtitles dataset. (A) is a performance comparison with respect to the number of words for adaptation. (B) is the number of iterations until the convergence during the finetuning stage with respect to the number of words. (C) is the number of iterations until convergence, where the BLEU is validating scores calculated by the average of En-De and De-En.}
    \label{fig:fig_2_3_4}
\vspace{-0.2in}
\end{figure*}
\subsection{Experimental Settings}
As our base model, we use a Transforme~\cite{vaswani2017attention} is initialized by a masked language model from XLM~\cite{conneau2019cross} using our out-domain datasets. All the models consist of six layers, 1,024 units, and eight heads. 


We establish and evaluate various baseline models as follows: \textbf{1. UNMT model} is only trained with in-domain monolingual data, composed of 5,000 words for each language. 
\textbf{2. Supervised neural machine translation model (NMT)} is trained with in-domain parallel datasets, 
which we arrange in parallel with the two in-domain monolingual corpora.
\noindent\textbf{3. Unadapted model} is pretrained with only the out-domain datasets and evaluated on the in-domain datasets. 
\textbf{4. Transfer learning model} is a finetuned model, which is pretrained with the out-domain datasets and then finetuned with a low-resource in-domain dataset. 
\rv{\textbf{5. Mixed finetuned model}~\cite{chu2017empirical} is similar to a Transfer learning model, but it utilizes both in-domain and out-domain datasets for finetuning. That is, the training batch is sampled evenly from in-domain and out-of-domain datasets.}
\subsection{Experimental Results}
%
In order to verify the effectiveness of leveraging the high-resource domains (i.e., source domains) to handle the low-resource domains (i.e., target domain), we compare the unsupervised and supervised models with ours and other baseline models. The unsupervised model is trained on in-domain data which significantly suffers from data scarcity in that it only uses low-resource in-domain data.

Although the unsupervised and supervised models are initialized by XLM, as shown in Table~\ref{table.main_result}, those models show the worst performance in all the cases. This result indicates that when the small size of an in-domain corpus is given, it is appropriate to utilize the out-domain datasets rather than to train only with low-resource data.
In addition, the performance of the unadapted model is far behind against other models, such as the mixed finetuned model, transfer learning model, \toolnameMeta, and \ourtoolname. This implies that we need an adequate strategy of leveraging the high-resource domains to improve the performance.

We further compare the performance between our proposed approaches (i.e., \toolnameMeta and \ourtoolname) and \rv{the other two finetuning models (i.e., the transfer learning and the mixed finetuned} models). Our methods exhibit the leading performances in both directions of translation ($en \leftrightarrow de$). They consistently achieve improvements of 2-4 BLEU score in most of the settings.
Furthermore, \ourtoolname consistently obtains better BLEU scores and converges faster than \toolnameMeta does. We assert that our proposed losses, i.e., the aggregated meta-train and the cross-domain losses, help the model to easily adjust to the unseen in-domain dataset, and thus accelerate the convergence speed.




\begin{table*}[t]
\centering
\resizebox{\textwidth}{!}{\begin{tabular}{l|rr|rr|rr|rr|rr|rr|rr|rr}
\hline
    \multicolumn{1}{c|}{Parameter}          & \multicolumn{12}{c|}{Initial $\theta$}                                                                                                                                                                                                                                                                                                                                                                                  & \multicolumn{4}{c}{Finetuned $\theta$}                                                                                                 \\ \hline
\multicolumn{1}{c|}{\multirow{2}{*}{D}} & \multicolumn{12}{c|}{$\mathcal{D}_{out}$}                                                                                                                                                                                                                                                                                                                                                                                                 & \multicolumn{2}{c|}{$\mathcal{D}_{in}$}                                            & \multicolumn{2}{c}{Unseen}                                       \\ \cline{2-17} 
\multicolumn{1}{c|}{}                   & \multicolumn{2}{c|}{Meidcal}                                      & \multicolumn{2}{c|}{Law}                                         & \multicolumn{2}{c|}{Koran}                                             & \multicolumn{2}{c|}{EUB}                                         & \multicolumn{2}{c|}{IT}                                          & \multicolumn{2}{c|}{GV}                                           & \multicolumn{2}{c|}{Subtitles}                                    & \multicolumn{2}{c}{Europarl}                                     \\ \hline
                   & De-En                           & En-De                           & De-En                           & En-De                          & De-En                                & En-De                           & De-En                           & En-De                          & De-En                          & En-De                           & De-En                           & En-De                           & De-En                           & En-De                           & De-En                           & En-De                           \\ \hline
Transfer                                & 30.98                           & 26.96                           & 34.8                            & 30.28                          & 13.72                                & 11.59                           & 12.32                           & 10.01                          & 20.98                          & 17.74                           & 17.4                            & 14.25                           & 10.92                           & 9.18                            & 22.31                           & 16.58                           \\
Mixed finetuned                          & -                           & -                   & -                         & -                         & -                                & -                         & -                    & -               & -                         & -                          & -                       & -                      & 11.77                           & 9.96                            & 22.84                           & 16.92                           \\
MetaUMT                                 & 33.0                     & 23.39                           & 27..74                          & 15.4                           & 4.89                                 & 0.79                            & 6.78                            & 2.59                           & 9.45                           & 4.68                            & 2.77                            & 1.06                            & 12.95                           & 10.58                           & 23.91                           & 18.7                            \\
MetaGUMT                                & \textbf{37.37} & \textbf{31.63} & \textbf{42.73} & \textbf{37.3} & \textbf{18.2} & \textbf{13.84} & \textbf{13.72} & \textbf{11.8} & \textbf{24.0} & \textbf{19.24} & \textbf{21.24} & \textbf{17.38} & \textbf{13.45} & \textbf{10.89} & \textbf{24.44} & \textbf{19.31} \\ \hline

\end{tabular}}
\vspace{-2mm}
\caption{\rv{BLEU scores evaluated on out-domain and in-domain data with initial $\theta$ and finetuned $\theta$, respectively. `$\mathcal{D}$' denotes the domain, `Unseen' indicates the new domain evaluated with finetuned $\theta$. Since the transfer and mixed finetuned model use the same initial $\theta$, we leave its corresponding row as `-'.}}
\vspace{-4mm}
\label{table.zero}
\end{table*}
\vspace{-2mm}

\subsection{Performances and Adaptation Speed in Finetuning Stage}


As shown in Fig.~\ref{fig:fig_2_3_4} (A), we compare our proposed methods with the transfer learning approach by varying the sizes of an in-domain monolingual corpus. The smaller training data is, the wider the gap between the performances of the two approaches and the transfer learning model becomes. It means that the meta-learning is an effective approach to alleviate the performance degradation, preventing the model from overfitting in the low-resource data. 


Compared to the transfer learning model, \toolnameMeta demonstrates a better performance than other methods in various settings. However, \ourtoolname exhibits even better performances consistently in all settings owing to our proposed losses (Eq.~\eqref{eqn:meta_test.3}).
The transfer learning approach shows the worst performance except for the unadapted model, even though it exploits the in-domain corpus after being pretrained with the out-domain datasets.


Additionally, we analyze the number of iterations required for a model to converge given an in-domain dataset. As shown in Fig.~\ref{fig:fig_2_3_4} (B), the meta-learning approaches rapidly converge after only a few iterations, even faster than the transfer learning one does. 
As the number of in-domain training words increases, the transfer learning approach requires a much larger number of iterations until convergence than our meta-learning approaches do. 
It can be seen that \toolnameMeta and \ourtoolname rapidly adapt to an unseen domain. Moreover, owing to the encapsulated knowledge from the high-resource domains, \ourtoolname converges within a relatively earlier iteration than \toolnameMeta does. 


In summary, the meta-learning-based methods quickly converge in the low-resource domain, improving the performances up to 2.2-4.1 BLEU score over the transfer learning method in various low-resource settings. This indicates that the meta-learning-based approaches are suitable to alleviate the data deficiency issue on scarce domains.
Furthermore, our proposed losses (Eq.~\eqref{eqn:meta_test.3}) enhance the capabilities of aggregating domain general knowledge and finding adequate initialization. 

\subsection{Number of Iterations until Convergence in Pretraining Stage}



An advantage of our meta-learning approaches is that they can find an optimal initialization point from which the model can quickly adapt to a low-resource in-domain dataset. The transfer learning model requires twice more iterations until convergence than ours do. As shown in Fig.~\ref{fig:fig_2_3_4} (C), \toolnameMeta and \ourtoolname not only converge quickly but also outperform the other baseline methods. Specifically, compared to \toolnameMeta, \ourtoolname is effective in achieving a proper initialization at an earlier iteration. These results indicate that our additional losses, i.e., the cross-domain and aggregated meta-train losses, are beneficial in boosting up the ability for finding an optimal initialization point when training the model with the out-domain datasets.

\subsection{Analysis of \ourtoolname losses }
\label{sec.5.7}

We assume that the domain generalization ability and high-resource domain knowledge is helpful for the UNMT model to translate the low-resource domain sentences. 
At first, to identify whether the model encapsulates the high-resource knowledge from multiple sources, we evaluate our model on out-domain datasets (i.e., $D_{out}$) with initial $\theta$. 
As shown in Table.~\ref{table.zero}, \ourtoolname shows remarkable performances than \toolnameMeta in all domains, even better than the transfer learning models. On the other hands, \toolnameMeta demonstrates poor performances in $D_{out}$. Compare to \toolnameMeta, \ourtoolname uses aggergated meta-train loss such that  
\ourtoolname is able to encapsulates the high-resource domain knowledge. 
As shown in Table.~\ref{table.main_result}, \ourtoolname shows the superior performances that \ourtoolname is capable to leverage the encapsulated knowledge when finetuning the low-resource target domain. Secondly, our cross-domain loss encourages the model to have generalization capability after adapting the low-resource target domain.
As `Unseen' column in Table.~\ref{table.zero}, \ourtoolname outperforms the other models. 
It can be seen that our model has the domain generalization ability after finetuning stage due to the cross-doamin loss in the meta-test phase.

\begin{table}[t]
\centering
\resizebox{\columnwidth}{!}{\begin{tabular}{c|c|rr|rr}
\hline
                                          Cross-domain& Aggregated meta-train& De-En & En-De & Average & $\Delta$ \\ 
\hline
                \xmark & \xmark & 27.09 & 24.6 & 25.85 & \\
\hline
                 \checkmark& \xmark & 27.37 & 24.76 & 26.06 & +0.21\\
                 \xmark & \checkmark & 27.54  & 24.9& 26.22& +0.37\\
                 \checkmark& \checkmark & \textbf{27.85} & \textbf{25.06} & \textbf{26.46}& +0.61  \\

\hline

\end{tabular}}

\caption{Effectiveness of each cross-domain and aggregated meta-train loss.}
\label{table.ablation_study}
\vspace{-0.22\in}
\end{table}

\subsection{Ablation Study}\label{ablation}
We empirically show the effectiveness of the cross-domain and aggregated meta-train losses, as shown in Table~\ref{table.ablation_study}~\footnote{The models are pretrained on Subtitles, Law, EUbookshop, Europarl, IT, and GlobalVoices datasets and then finetuned on a Medical dataset.}. First, compared to \toolnameMeta that does not use any of the two losses, incorporating the cross-domain loss improves the average BLEU score by 0.21. The cross-domain loss acts as a regularization function that prevents the model to overfit during the finetuning stage. Second, the aggregated meta-train loss, another critical component of our model, allows the model to utilize the high-resource domain knowledge in the finetuning stage. This also improves the average BLEU score by 0.37 from \toolnameMeta. Lastly, combining both cross-domain and aggregated meta-train losses significantly enhance the result in both directions of translation ($En \leftrightarrow De$), indicating that they are complementary to each other.

\section{Conclusions}
This paper proposes novel meta-learning approaches for low-resource UNMT, called \toolnameMeta, which leverages multiple source domains to quickly and effectively adapt the model to the target domain even with a small amount of training data. Moreover, we introduce an improved method called \ourtoolname, which enhances cross-domain generalization and maintains high-resource domain knowledge. We empirically show that our proposed approach consistently outperforms the baseline methods with a nontrivial margin.

\bibliographystyle{acl_natbib}
\bibliography{anthology,acl2021}
\clearpage
\newpage
\clearpage
\appendix
\counterwithin{figure}{section}
\setcounter{table}{0}
\renewcommand{\thetable}{T.\arabic{table}}
\renewcommand{\thealgorithm}{A.\arabic{algorithm}}

\begin{table*}[tb]
\centering
\resizebox{\textwidth}{!}{\begin{tabular}{l|r|rrr|rrr|rrr}
\hline
\multirow{4}{*}{Model}   &
    \multirow{2}{*}{$\mathcal{D}_{out}$} 
                    & Medical & Law & Koran & Medical & Law & Koran & GV & Europarl &EUB \\
                    & & Subtitles & EUB & Europarl &  Subtitles & EUB & Europarl & Subtitles& Medical& Koran \\
                    
                    \cline{2-11}
    & \multirow{2}{*}{$\mathcal{D}_{in}$} 
                    & \multicolumn{3}{c|}{IT} & \multicolumn{3}{c|}{GV} &  \multicolumn{3}{c}{IT}\\\cline{3-11}
                    & &  De-En & En-De & epoch & De-En & En-De & epoch & De-En & En-De & epoch   \\
                    \hline
\multicolumn{2}{l|}{Unadapted}  &  18.62 & 14.89 & - & 19.27& 16.65 & - &   16.1& 15.3 & - \\
\multicolumn{2}{l|}{Transfer} & 19.80 & 16.35& 4  & 19.99 & 16.9& 3  & 19.31 & 16.13 & 5  \\ 
\multicolumn{2}{l|}{Mixed} & 19.75 & 16.49& 7  & 20.03 & 16.95 & 5  & 19.39 & 16.18 & 8  \\ 
\multicolumn{2}{l|}{{\toolnameMeta}} & 21.08& 18.05 & 4 & 22.36& 18.91&3& 20.5& 17.06 & \textbf{4}   \\ 
\multicolumn{2}{l|}{{\ourtoolname}} &  \textbf{21.37} &\textbf{18.42} & \textbf{3} &  \textbf{22.76}& \textbf{19.24}& \textbf{2}& \textbf{20.74}  & \textbf{17.74} &\textbf{4}\\ 
\hline    
\multicolumn{2}{l|}{Supervised NMT} & 3.48 & 3.33 & 15 & 0.97 & 0.85 & 14 & 3.53 & 3.59 & 10 \\
\multicolumn{2}{l|}{Unsupervised NMT} & 1.83 & 0.86 & 22 & 0.51 & 0.18 & 20  & 0.51 & 0.55 & 7 \\
\hline
\end{tabular}}

\caption{
Extended results on various domain settings.
The column `epoch' indicates the converged number of epochs for each in-domain dataset. Since the unadapted model does not involve an additional finetuning step, we leave the epoch column as blank.}
\label{table.main_result_2}
\end{table*}
\begin{table}[]
\resizebox{\columnwidth}{!}{\begin{tabular}{l|ll|ll|ll|ll}
\hline
\multicolumn{1}{c|}{\multirow{2}{*}{$\mathcal{D}_{out}$}} & \multicolumn{2}{c|}{MetaGUMT}   & \multicolumn{2}{c|}{MetaUMT} & \multicolumn{2}{c|}{Transfer} & \multicolumn{2}{c}{Mixed}\\ \cline{2-9}
\multicolumn{1}{c|}{}                         & en-de          & de-en         & en-de        & de-en        & en-de         & de-en & en-de         & de-en    \\ \hline
Medical-Law-Koran-IT                          & \textbf{5.97}  & \textbf{7.47} & 5.87         & 7.24         & 5.75          & 7.17  & 5.87 & 7.22       \\
Medical-Law-Koran-IT-GV                       & \textbf{7.58}  & \textbf{9.49} & 7.33         & 9.01         & 7.17          & 8.08     & 7.20 & 8.68\\
Medical-Law-Koran-IT-GV-EUB                      & \textbf{10.89} & \textbf{13.45}          & 10.58       & 12.95       & 9.18          & 10.92 & 9.96 & 11.77       \\ \hline
\end{tabular}}
\caption{Effectiveness of the different number of source domains between meta-learning based approaches and the transfer learning approach.}
\label{table.num_domains}

\end{table}
\section{Implementation Details}
In order to preprocess datasets, We utilize Moses~\cite{koehn2007moses} to tokenize the sentences. 
We then use byte-pair encoding (BPE)~\cite{sennrich2016improving} to build a shared sub-word vocabulary using fastBPE\footnote{https://github.com/glample/fastBPE} with 60,000 BPE codes. 
Based on this shared sub-word vocabulary, constructed from the out-domain datasets, we split words into sub-word units for the in-domain dataset.
\rv{We implement all of the models using PyTorch library~\footnote{\url{https://pytorch.org/}}, and then train them in four nvidia V100 gpus for pretraining and finetuning. 
We evaluate all the experiments based on the BLEU script~\footnote{\url{https://github.com/moses-smt/mosesdecoder/blob/master/scripts/generic/multi-bleu.perl}}. The number of convergence iteration of each algorithm is defined based on the best validation epoch, which shows no more improvement of validation score even if we run 10 more epochs. 
Moreover, we have conducted comprehensive experiments to obtain our main result table (Table.~\ref{table.main_result} and Table.~\ref{table.main_result_2} ) on different domains by training the model with 10 different sampled words each time.
}

\rv{
For optimizing each algorithms, We choose the Adam optimizer~\cite{kingma2014adam} for pretraining stage, as well as the Adam warmup optimizer~\cite{vaswani2017attention} for finetuning stage. The learning rate is set to $10^{-4}$,  optimized within the range of $10^{-2}$ to $10^{-5}$. In all experiments, the number of tokens per batch is set as 1,120 and the dropout rate is set as 0.1. In meta-learning approaches, we set the learning rates of alpha and beta commonly as 0.0001 in all experiments. }

\section{Additional results}
\rv{Due to the page limits, in this section, we provide a few additional results that include other domain combinations shown in Table.~\ref{table.main_result_2}. Clearly, our proposed approaches still significantly outperform other baseline models on various domain settings.}

\subsection{Impact of the Number of Source Domains}
We examine how the performances change against the different number of source domains for each approach.
As shown in Table.~\ref{table.num_domains}, \ourtoolname consistently outperforms the transfer, the mixed-finetune, and \toolnameMeta approaches. 
As the size of the source domains increases, so does the performance gap between ours and the transferring based models, transferring and mixed-finetune models.
This indicates that the meta-learning based approaches are highly affected by the size of domains in the meta-train phase, and also, \rv{if the number of source domains are large enough to capture the general knowledge, the meta-learning based approaches are suitable to handle the low-resource target task, i.e., domain.}

\subsection{Perfomances and Adaptation Speed in Finetuning Stage for a Law domain}
\begin{figure}[]
    \centering
    \includegraphics[width=1\columnwidth]{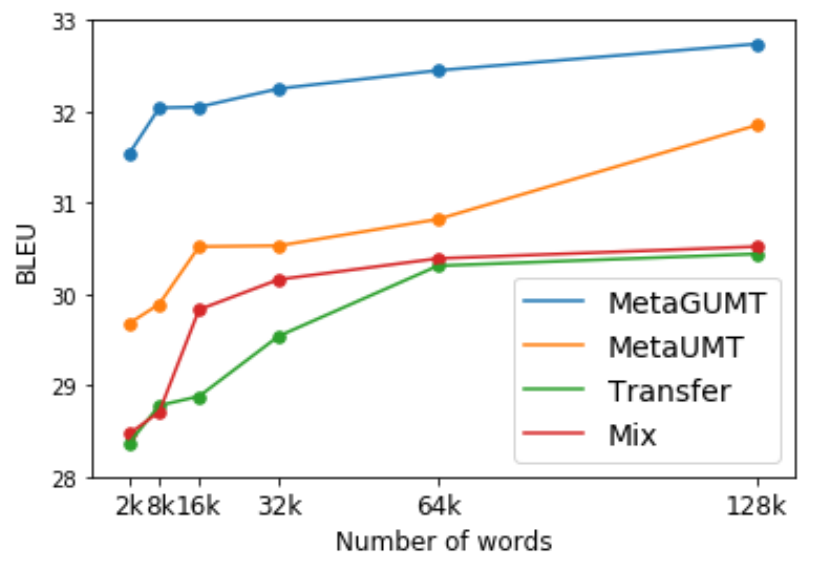}
    \caption{A performance comparison with respect to the number of words for adaptation on a Law domain.} 
    \label{fig:law_numsamples_performance}
\end{figure}

\begin{figure}[]
    \centering
    \includegraphics[width=1\columnwidth]{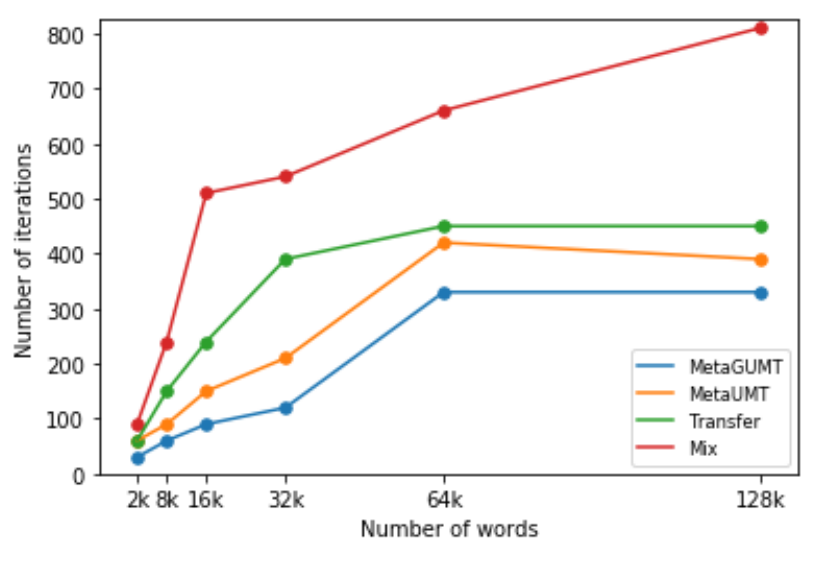}
    \caption{Number of iterations until the convergence during the finetuining stage with respect to the number of words on a Law domain.} 
    \label{fig:law_numsamples_epoch}
\end{figure}

As shown in Fig~\ref{fig:law_numsamples_performance} and Fig~\ref{fig:law_numsamples_epoch}, MetaGUMT consistently outperforms other methods even though the number words are increasing. Through this experiment, we attempt to show the robustness of our methods, MetaUMT and MetaGUMT, against others, transferring and mixed-finetune models. The models are pretrained on Subtitles, EUbookshop, Europarl, GlobalVoices, Medical, and Koran datasets and then finetuned on a Law dataset.
\

\section{Comparison between MetaGUMT and MetaUMT algorithms}
Here, we provide additional details of \toolnameMeta for understating the difference between the \ourtoolname and \toolnameMeta. We describe the overall algorithms of \toolnameMeta (\ref{alg:maml}) and \ourtoolname (\ref{alg:our_maml}). As shown in Algorithms.~\ref{alg:maml}, \toolnameMeta has only difference in line 10.

\begin{algorithm}[t]
\caption{\ourtoolname}
\label{alg:our_maml}
\textbf{Require}: $\alpha$, $\beta$: step sizes
\begin{algorithmic}[1] 
\STATE Pretrain $\theta$ by using XLM
\WHILE{not done}
\FOR{all $\mathcal{D}_{out}^{i}$}
\STATE Evaluate $\mathcal{\nabla}_{\theta} L^{\text{lm}}_{{\mathcal{D}}_{out}^{i}}(\theta)$ with respect to source and target language sentences from $\mathcal{D}_{out}^{i}$ 
\STATE \textbf{Back-translation} generates source and target language sentences using the current translation model
\STATE Evaluate $\nabla_{\theta}\mathcal{L}^{\text{bt}}_{{\mathcal{D}}_{out}^{i}}(\theta)$ with using pseudo-generated sentences
\STATE Sum each gradient: \\ $\nabla_{\theta}\mathcal{L}^{s}_{\mathcal{D}_{out}^{i}} = \nabla_{\theta}\mathcal{L} ^{\text{lm}}_{{\mathcal{D}}_{out}^{i}}(\theta)+ \nabla_{\theta}\mathcal{L}^{\text{bt}}_{{\mathcal{D}}_{out}^{i}}(\theta)$
\STATE Compute adapted parameters with one-step gradient descent:\\ $\phi^{i} = \theta - \alpha\nabla_{\theta}\mathcal{L}^{\text{s}}_{D_{out}^{i}}(\theta)$
\ENDFOR
\STATE Update $\theta \leftarrow \theta - \beta\nabla_{\theta}(\mathcal{L}_{cd} + \mathcal{L}_{ag})$
\ENDWHILE
\end{algorithmic}
\end{algorithm}
\begin{algorithm}[]
\caption{\toolnameMeta}
\label{alg:maml}
\textbf{Require}: $\alpha,\beta$: step sizes
\begin{algorithmic}[1] 
\STATE Pretrain $\theta$ by using XLM
\WHILE{not done}
\FOR{all $\mathcal{D}_{out}^{i}$}
\STATE Evaluate $\mathcal{\nabla_{\theta} L}^{\text{lm}}_{{\mathcal{D}}_{out}^{i}}(\theta)$ with respect to source and target language sentences from $\mathcal{D}_{out}^{i}$
\STATE \textbf{Back-translation} generates source and target language sentences using the current translation model
\STATE Evaluate $\nabla_{\theta}\mathcal{L}^{\text{bt}}_{\mathcal{D}_{out}^{i}}(\theta)$ with using pseudo-generated sentences
\STATE Sum each gradient:\\$\nabla_{\theta}\mathcal{L}^{s}_{\mathcal{D}_{out}^{i}} = \nabla_{\theta}\mathcal{L}^{\text{lm}}_{\mathcal{D}_{out}^{i}}(\theta)+ \nabla_{\theta}\mathcal{L}^{\text{bt}}_{\mathcal{D}_{out}^{i}}(\theta)$ 
\STATE Compute adapted parameters with one-step gradient descent: \\ $\phi^{i} = \theta - \alpha\nabla_{\theta}\mathcal{L}_{D_{out}^{i}}^{s}(\theta)$
\ENDFOR
\STATE $\theta \leftarrow \theta-\beta\nabla_{\theta}{\sum}^{n}_{i=0}\mathcal{L}^{\text{s}}_{\mathcal{D}_{out}^{i}}(\phi^{i})$
\ENDWHILE
\end{algorithmic}
\end{algorithm}

\section{Statistics of Datasets}

As shown in Table.~\ref{table.dataset}, W/S indicates the number of words per sentence in a domain.
\label{sec:appendix}
\begin{table}[t]
\centering
\resizebox{\columnwidth}{!}{\begin{tabular}{l|c|c|c|c|c}
\hline
                                          \multirow{2}{*}{Corpus}& \multicolumn{2}{c|}{Words} & \multirow{2}{*}{Sentences} & \multicolumn{2}{c}{W/S}  \\ \cline{2-3}\cline{5-6}
& EN&DE & & EN & DE \\
\cline{1-6}
Acquis (Law) & 9.2M & 8M & 0.7M & 12.93&11.30\\
EMEA (Medical) & 7.5M & 6.3M & 1.1M & 6.81 & 5.75\\
IT & 1.7M & 1M & 0.3M & 9.08 & 5.32 \\
Tanzil (Koran) & 5.6M & 5.3MS & 0.5M & 10.66 & 10.08 \\
Subtitles  & 92.7M & 87.6M & 22.5M & 4.11 & 3.89 \\
EUbookshop (EUB) & 115.4M & 100M & 9.3M & 12.37 & 10.72 \\
Europarl & 27.3M & 25.7M & 1.9M & 13.99 & 13.18 \\
GlobalVoices (GV) & 0.6M & 0.6M & 0.05M & 10.67 & 10.88 \\
 
\hline
\end{tabular}}
\caption{Statistics of each corpora.}
\label{table.dataset}
\end{table}

\end{document}